\documentclass[letterpaper,twocolumn,fleqn]{article} 

\usepackage{array}
\newcolumntype{x}[1]{>{\centering\arraybackslash}p{#1}}

\usepackage{caption}
\usepackage{subcaption}
\usepackage{bm}
\usepackage{booktabs}
\usepackage{multirow}

\usepackage{arydshln}
\usepackage{amsmath}
\usepackage{amssymb}
\usepackage{pifont}
\newcommand{\cmark}{\ding{51}}
\newcommand{\xmark}{\ding{55}}

\usepackage[svgnames]{xcolor}

\usepackage{tikz}
\usepackage{tikz-cd}
\usetikzlibrary{shapes, arrows, positioning, decorations.pathreplacing, calc, external}

\usepackage[hidelinks]{hyperref}

\usepackage{ist}
\def\A{\bm{\mathrm{A}}}
\def\B{\bm{\mathrm{B}}}

\title{A deep perceptual metric for 3D point clouds}
\author{Maurice Quach\textsuperscript{†}, Aladine Chetouani\textsuperscript{†+}, Giuseppe Valenzise\textsuperscript{†}, Frederic Dufaux\textsuperscript{†};\newline \textsuperscript{†} Universit\'e Paris-Saclay, CNRS, CentraleSup\'elec, Laboratoire des signaux et syst\`emes; 91190 Gif-sur-Yvette, France \newline \textsuperscript{+} Laboratoire PRISME, Université d’Orl\'eans; Orl\'eans, France}

\date{}
\begin{document} 

\maketitle 
\thispagestyle{empty}

\begin{abstract}
Point clouds are essential for storage and transmission of 3D content. As they can entail significant volumes of data, point cloud compression is crucial for practical usage. Recently, point cloud geometry compression approaches based on deep neural networks have been explored. In this paper, we evaluate the ability to predict perceptual quality of typical  voxel-based loss functions employed to train these networks. We find that the commonly used focal loss and weighted binary cross entropy are poorly correlated with human perception. We thus propose a perceptual loss function for 3D point clouds which outperforms existing loss functions on the ICIP2020 subjective dataset.
In addition, we propose a novel truncated distance field voxel grid representation and find that it leads to sparser latent spaces and loss functions that are more correlated with perceived visual quality compared to a binary representation.
The source code is available at \url{https://github.com/mauriceqch/2021_pc_perceptual_loss}.
\end{abstract}

\section{Introduction}

As 3D capture devices become more accurate and accessible, point clouds are a crucial data structure for storage and transmission of 3D data. Naturally, this comes with significant volumes of data. Thus, Point Cloud Compression (PCC) is an essential research topic to enable practical usage.
The Moving Picture Experts Group (MPEG) is working on two PCC standards \cite{schwarz_emerging_2018}: Geometry-based PCC (G-PCC) and Video-based PCC (V-PCC).
G-PCC uses native 3D data structures to compress point clouds, while V-PCC employs a projection-based approach using video coding technology \cite{noauthor_itu-t_2019}.
These two approaches are complementary as V-PCC specializes on dense point clouds, while G-PCC is a more general approach suited for sparse point clouds.
Recently, JPEG Pleno \cite{ebrahimi_jpeg_2016} has issued a Call for Evidence on PCC \cite{noauthor_final_2020}.

Deep learning approaches have been employed to compress geometry \cite{quach_learning_2019, quach_improved_2020, wang_learned_2019, wang_multiscale_2020, tang_deep_2020, milani_syndrome-based_2020} and attributes \cite{quach_folding-based_2020,alexiou_towards_2020} of points clouds.
Specific approaches have also been developed for sparse LIDAR point clouds \cite{huang_octsqueeze_2020,biswas_muscle_2020}.
In this work, we focus on lossy point cloud geometry compression for dense point clouds.
In existing approaches, different point cloud geometry representations are considered for compression: G-PCC adopts a point representation, V-PCC uses a projection or image-based representation and deep learning approaches commonly employ a voxel grid representation.
Point clouds can be represented in different ways in the voxel grid. Indeed, voxel grid representations include binary and Truncated Signed Distance Fields (TSDF) representations \cite{curless_volumetric_1996}.
TDSFs rely on the computation of normals; however, in the case of point clouds this computation can be noisy. We then ignore the normal signs and reformulate TDSFs to propose a new Truncated Distance Field (TDF) representation for point clouds

Deep learning approaches for lossy geometry compression typically jointly optimize rate and distortion.
As a result, an objective quality metric, employed as a loss function, is necessary to define the distortion objective during training.
Such metrics should be differentiable, defined on the voxel grid and well correlated with perceived visual quality.
In this context, the Weighted Binary Cross Entropy (WBCE) and the focal loss \cite{lin_focal_2017} are commonly used loss functions based on a binary voxel grid representation.
They aim to alleviate the class imbalance between empty and occupied voxels caused by point cloud sparsity.
However, they are poorly correlated with human perception as they only compute a voxel-wise error.

\def\dstA{0.8cm}
\def\dstB{0.0cm}
\def\dstC{0.2cm}
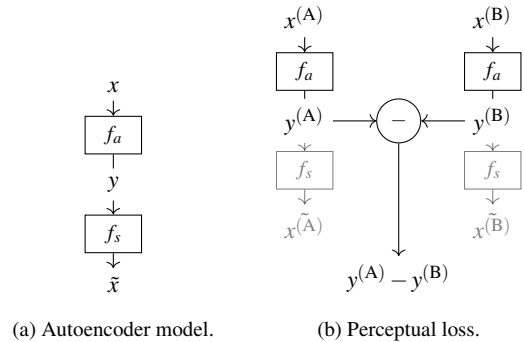
\begin{figure}[t]
\centering
\begin{subfigure}[b]{0.45\columnwidth}
	\centering
    \begin{tikzpicture}[
    		mainnode/.style={draw, rectangle, minimum width=0.75cm, minimum height=0.5cm, align=center,
    			font=\footnotesize},
            mininode/.style={rectangle, fill=white},
        ]
    	\node (in) {$x$};
    	\node (fa) [mainnode, below=\dstC of in.south, anchor=north] {$f_a$};
    	\node (fs) [mainnode, below=\dstA of fa.south, anchor=north] {$f_s$};
    	\node (out) [below=\dstC of fs.south, anchor=north, minimum height=0.4cm] {$\tilde{x}$};
    	\node (y) [mininode] at ($(fa)!0.5!(fs)$) {$y$};
    
    	\draw [->] (in) -- (fa);
    	\draw [-] (fa) -- (y);
    	\draw [->] (y) -- (fs);
        \draw [->] (fs) -- (out);
    \end{tikzpicture}
    \caption{Autoencoder model.}
    \label{fig:perceptual_loss_model}
\end{subfigure}
\begin{subfigure}[b]{0.45\columnwidth}
	\centering
    \begin{tikzpicture}[
    		mainnode/.style={draw, rectangle, minimum width=0.75cm, minimum height=0.5cm, align=center,
    			font=\footnotesize},
            mininode/.style={rectangle, fill=white},
        ]
    	\node (in) {$x^{(\A)}$};
    	\node (fa) [mainnode, below=\dstC of in.south, anchor=north] {$f_a$};
    	\node (fs) [mainnode, below=\dstA of fa.south, anchor=north, opacity=0.5] {$f_s$};
    	\node (out) [below=\dstC of fs.south, anchor=north, minimum height=0.4cm, opacity=0.5] {$\tilde{x^{(\A)}}$};
    	\node (y) [mininode] at ($(fa)!0.5!(fs)$) {$y^{(\A)}$};
    
    	\draw [->] (in) -- (fa);
    	\draw [-] (fa) -- (y);
    	\draw [->, opacity=0.5] (y) -- (fs);
        \draw [->, opacity=0.5] (fs) -- (out);
    	
    	\node (in2) [right=2.5cm of in.center, anchor=center] {$x^{(\B)}$};
    	\node (fa2) [mainnode, right=2.5cm of fa.center, anchor=center] {$f_a$};
    	\node (fs2) [mainnode, right=2.5cm of fs.center, anchor=center, opacity=0.5] {$f_s$};
    	\node (out2) [right=2.5cm of out.center, anchor=center, minimum height=0.4cm, opacity=0.5] {$\tilde{x^{(\B)}}$};
    	\node (y2) [mininode] at ($(fa2)!0.5!(fs2)$) {$y^{(\B)}$};
    
    	\draw [->] (in2) -- (fa2);
    	\draw [-] (fa2) -- (y2);
    	\draw [->, opacity=0.5] (y2) -- (fs2);
        \draw [->, opacity=0.5] (fs2) -- (out2);
        
        \node (ydiff) [draw, circle, align=center, font=\footnotesize] at
        ($(y)!0.5!(y2)$) {$-$};
        \draw [->] (y) -- (ydiff);
        \draw [->] (y2) -- (ydiff);
        
        \node (ydiffvar) [below=1.5cm of ydiff] {$y^{(\A)} - y^{(\B)}$};
        \draw [->] (ydiff) -- (ydiffvar);
    \end{tikzpicture}
    \caption{Perceptual loss.}
    \label{fig:perceptual_loss_loss}
\end{subfigure}
\caption{Perceptual loss based on an autoencoder. The grayed out parts do not need to be computed for the perceptual loss.}
\end{figure}

\begin{figure*}[t]
  \centering
\begin{subfigure}[b]{0.33\textwidth}
	\centering
    \includegraphics[width=\linewidth, height=5cm]{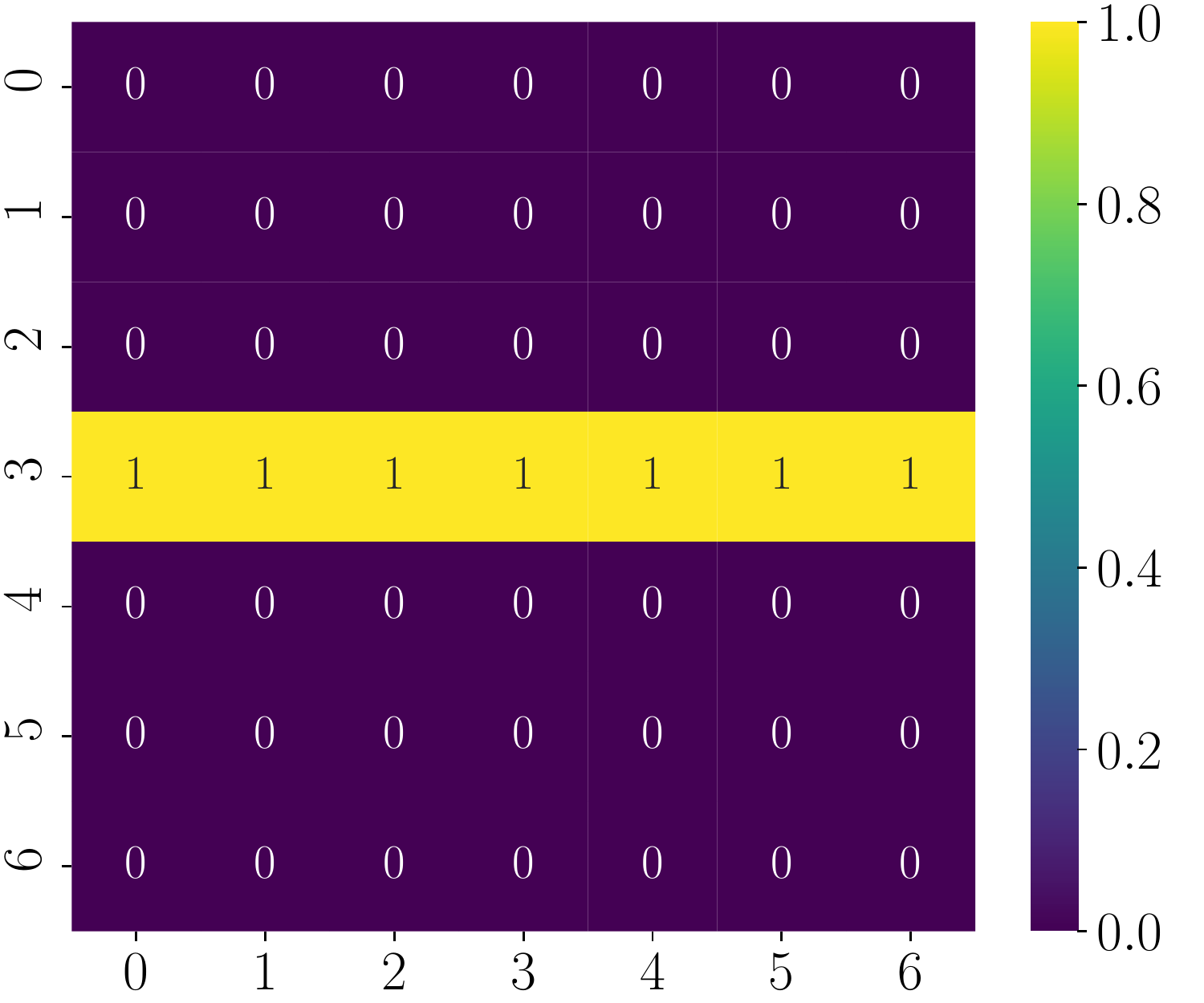}
    \caption{Binary.}
    \label{fig:voxel_grid_repr_bin}
\end{subfigure}
\begin{subfigure}[b]{0.33\textwidth}
	\centering
    \includegraphics[width=\linewidth, height=4.9cm]{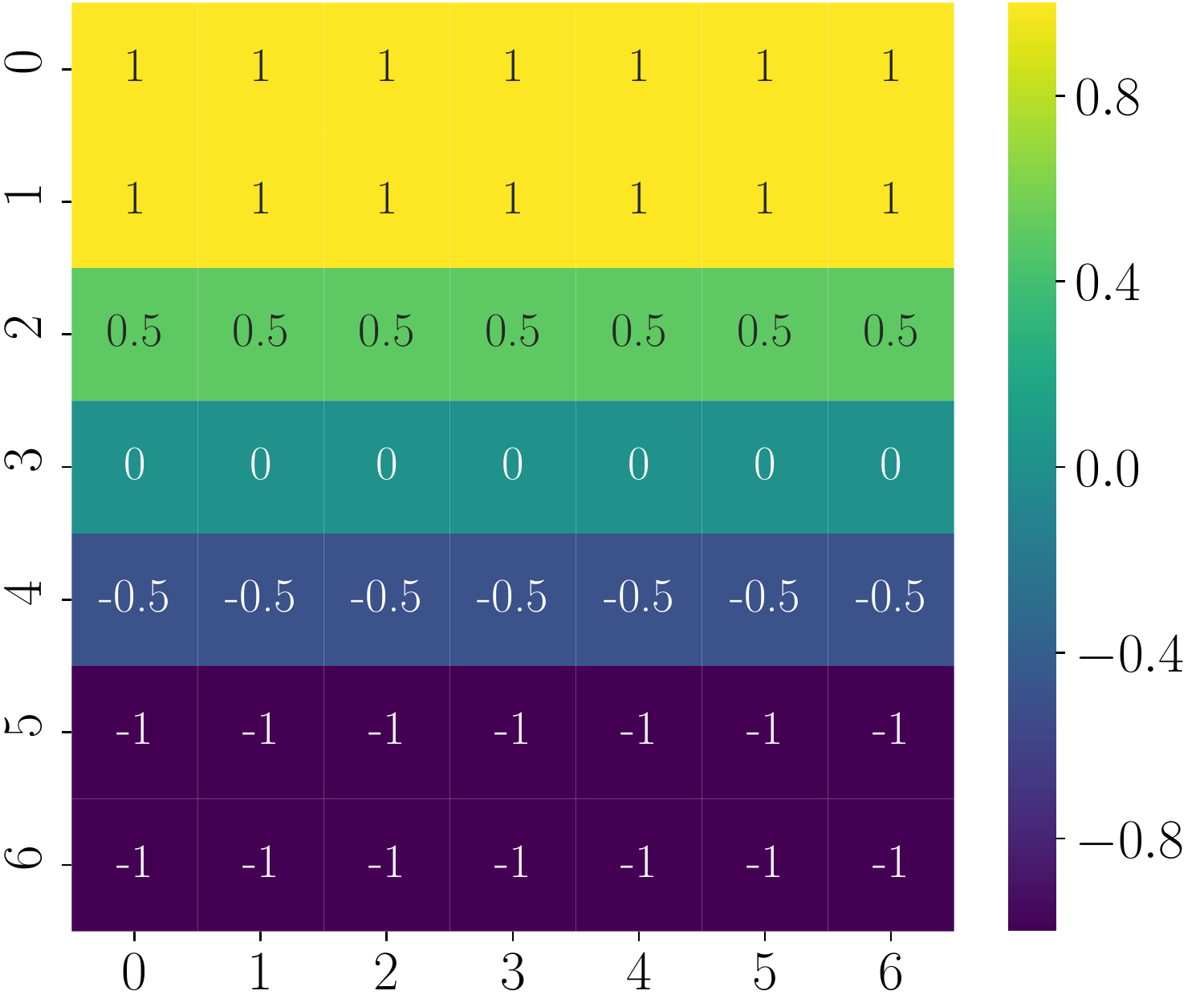}
    \caption{Truncated Signed Distance Field (TSDF).}
    \label{fig:voxel_grid_repr_tsdf}
\end{subfigure}
\begin{subfigure}[b]{0.33\textwidth}
	\centering
    \includegraphics[width=\linewidth, height=5cm]{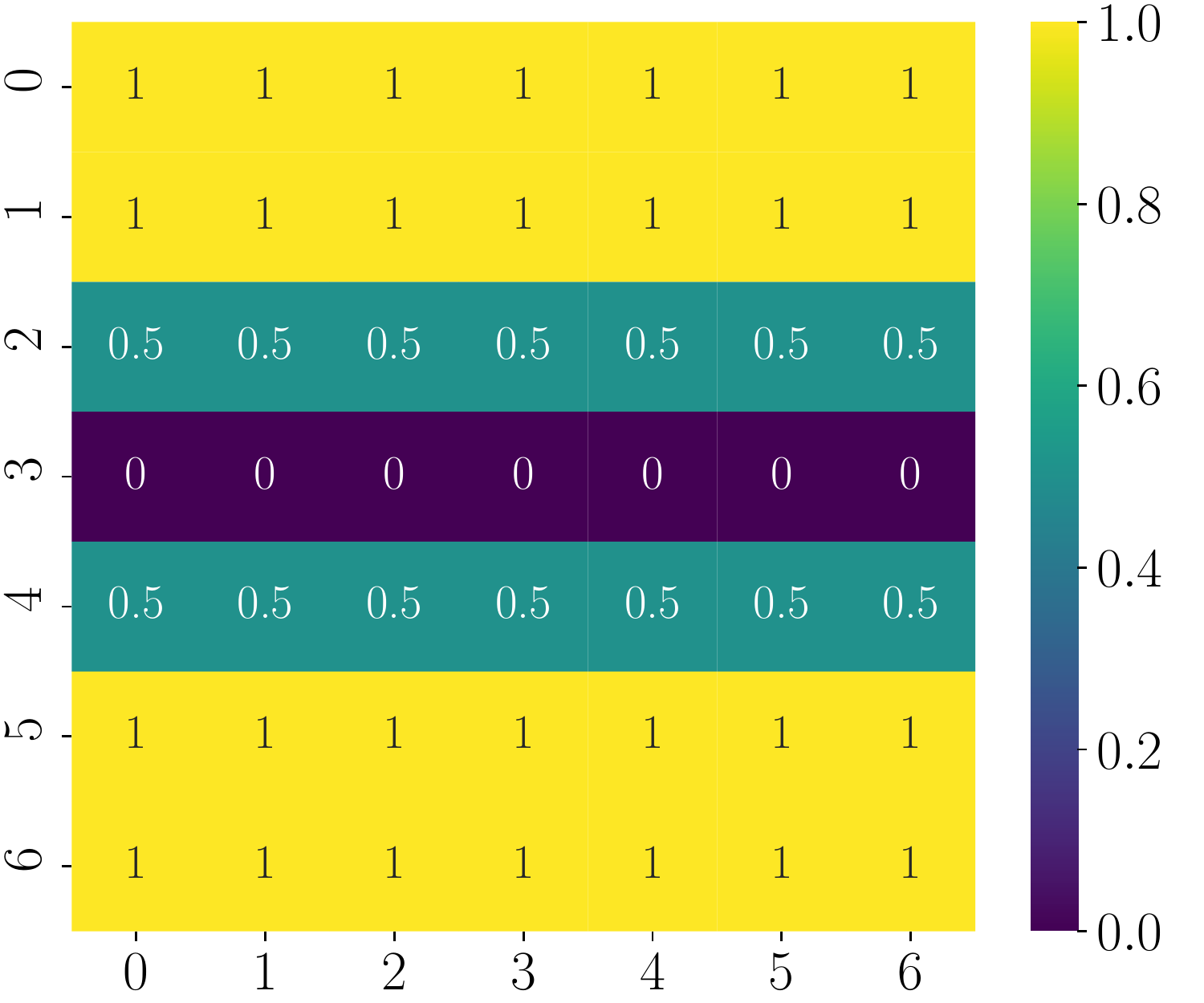}
    \caption{Truncated Distance Field (TDF).}
    \label{fig:voxel_grid_repr_tdf}
\end{subfigure}
  \caption{Voxel grid representations of a point cloud. The upper bound distance value is 2 for the TDF and the TSDF. Normals are facing up in the TSDF.}
  \label{fig:voxel_grid_repr}
\end{figure*}

\begin{table*}[ht]
\centering
\caption{Table \ref{tab:metrics_table}: Objective quality metrics considered in this study.}
\label{tab:metrics_table}
\begin{tabular}{@{}p{1cm}lp{1.75cm}x{1.75cm}x{1.5cm}p{7cm}@{}} 
\toprule
Domain & Name & Signal type & Block aggregation & Learning based & Description \\ \midrule
\multirow{4}{*}{Points} & D1 MSE & Coordinates & \textcolor{Red}{\xmark} & \textcolor{Red}{\xmark} & Point-to-point MSE \\ \cdashline{2-6}
& D2 MSE & Coordinates & \textcolor{Red}{\xmark} & \textcolor{Red}{\xmark} & Point-to-plane MSE \\ \cdashline{2-6}
& D1 PSNR & Coordinates & \textcolor{Red}{\xmark} & \textcolor{Red}{\xmark} & Point-to-point PSNR \\ \cdashline{2-6}
& D2 PSNR & Coordinates & \textcolor{Red}{\xmark} & \textcolor{Red}{\xmark} & Point-to-plane PSNR \\ \midrule
\multirow{8}{*}{\shortstack[l]{Voxel\\ Grid}} & Bin BCE & Binary & L1 & \textcolor{Red}{\xmark} & Binary cross entropy \\ \cdashline{2-6}
& Bin naBCE & Binary & L1 & \textcolor{Red}{\xmark} & Neighborhood adaptive binary cross entropy \cite{guarda_neighborhood_2020} \\ \cdashline{2-6}
& Bin WBCE 0.75 & Binary & L2 & \textcolor{Red}{\xmark} & Weighted binary cross entropy with $w = 0.75$ \\ \cdashline{2-6}
& Bin PL & Binary & L1 & \textcolor{Green}{\cmark} & Perceptual loss (explicit) on all feature maps \\ \cdashline{2-6}
& Bin PL F1 & Binary & L1 & \textcolor{Green}{\cmark} & Perceptual loss (explicit) on feature map 1 \\ \cdashline{2-6}
& TDF MSE & Distances & L1 & \textcolor{Red}{\xmark} & Truncated distance field (TDF) MSE \\ \cdashline{2-6}
& TDF PL & Distances & L1 & \textcolor{Green}{\cmark} & Perceptual loss (implicit) over all feature maps \\ \cdashline{2-6}
& TDF PL F9 & Distances & L1 & \textcolor{Green}{\cmark} & Perceptual loss (implicit) on feature map 9 \\ \cdashline{2-6}
\bottomrule
\end{tabular}
\end{table*}

A number of metrics have been proposed for Point Cloud Quality Assessment (PCQA): the point-to-plane (D2) metric \cite{tian_geometric_2017}, PC-MSDM \cite{meynet_pc-msdm:_2019}, PCQM \cite{meynet_pcqm_2020}, angular similarity \cite{alexiou_point_2018}, point to distribution metric \cite{javaheri_mahalanobis_2020}, point cloud similarity metric \cite{alexiou_towards_2020-1}, improved PSNR metrics \cite{javaheri_improving_2020} and a color based metric \cite{viola_color-based_2020}. These metrics operate directly on the point cloud.
However, they are not defined on the voxel grid and hence cannot be used easily as loss functions.
Recently, to improve upon existing loss functions such as the WBCE and the focal loss, a neighborhood adaptive loss function \cite{guarda_neighborhood_2020} was proposed.
Still, these loss functions are based on the explicit binary voxel grid representation.
We show in this paper that loss functions based on the TDF representation are more correlated with human perception than those based on the binary representation.

The perceptual loss has previously been proposed as an objective quality metric for images \cite{zhang_unreasonable_2018}.
Indeed, neural networks learn representations of images that are well correlated with perceived visual quality.
This enables the definition of the perceptual loss as a distance between latent space representations.
For the case of images, the perceptual loss provides competitive performance or even outperforms traditional quality metrics.
We hypothesize that a similar phenomenon can be observed for point clouds.

Therefore, we propose a differentiable perceptual loss for training deep neural networks aimed at compressing point cloud geometry.
We investigate how to build and train such a perceptual loss to improve point cloud compression results.
Specifically, we build a differentiable distortion metric suitable for training neural networks to improve PCC approaches based on deep learning. We then validate our approach experimentally on the ICIP2020 \cite{perry_quality_2020} subjective dataset.
The main contributions of the paper are as follows:
\begin{itemize}
    \item A novel perceptual loss for 3D point clouds that outperforms existing metrics on the ICIP2020 subjective dataset
    \item A novel implicit TDF voxel grid representation
    \item An evaluation of binary (explicit) and TDF (implicit) representations in the context of deep learning approaches for point cloud geometry compression
\end{itemize}

\section{Voxel grid representations}

In this study, we consider different voxel grid representations for point clouds.
A commonly used voxel grid representation is the explicit binary occupancy representation where the occupancy of a voxel (occupied or empty) is represented with a binary value (Figure \ref{fig:voxel_grid_repr_bin}).
In the binary (Bin) representation (Figure \ref{fig:voxel_grid_repr_bin}), each voxel has a binary occupancy value indicating whether it is occupied or empty. When the $i$th voxel is occupied, then $x_i = 1$ and otherwise $x_i = 0$.

Another representation is the implicit TSDF representation which has been employed for volume compression \cite{tang_deep_2020}.
Instead of an occupancy value, the value of a voxel is the distance from this voxel to the nearest point and the sign of this value is determined from the orientation of the normal (Figure \ref{fig:voxel_grid_repr_tsdf}).
However, this requires reliable normals which may not be available in sparse and/or heavily compressed point clouds.

Hence, we propose an implicit TDF representation which is a variant of the TSDF without signs and therefore does not require normals.
In the implicit TDF representation (Figure \ref{fig:voxel_grid_repr_tdf}), the $i$th voxel value is $x_i = d$, where $d$ is the distance to its nearest occupied voxel. Consequently, $x_i = 0$ when a voxel is occupied and $x_i = d$ with $d > 0$ otherwise.
Additionally, we truncate and normalize the distance values into the $[0, 1]$ interval with
\begin{equation}
    x_i = \min(d, u) / u,
\label{eq:tdf_trunc_norm}
\end{equation}
where $u$ is an upper bound value.

In this study, we focus on the explicit binary and implicit TDF representations.

\section{Objective quality metrics}

In Table \ref{tab:metrics_table}, we present the objective quality metrics considered in this study. Specifically, we evaluate metrics that are differentiable on the voxel grid to evaluate their suitability as loss functions for point cloud geometry compression.
We include metrics defined on binary and TDF representations and we compare their performance against traditional point set metrics.

\subsection{Voxel grid metrics}

We partition the point cloud into blocks and compute voxel grid metrics for each block. For each metric, we aggregate metric values over all blocks with either L1 or L2 norm. Specifically, we select the best aggregation experimentally for each metric.

Given two point clouds $\A$ and $\B$, we denote the $i$th voxel value for each point cloud as $x^{(\A)}_i$ and $x^{(\B)}_i$.
Then, we define the WBCE as follows
\begin{align}
\begin{split}
    -\frac{1}{N} \sum_{i} \Bigl( &\alpha x^{(\A)}_{i}\log(x^{(\B)}_{i})\\
    &+ (1 - \alpha)(1 - x^{(\A)}_{i})\log(1 - x^{(\B)}_{i}) \Bigr),
\end{split}
\end{align}
where $\alpha$ is a balancing weight between $0$ and $1$. The binary cross entropy (BCE) refers to the case $\alpha = 0.5$.

Different from the WBCE, the Focal Loss (FL) amplifies ($\gamma > 1$) or reduces ($\gamma < 1$) errors and is defined as follows
\begin{align}
\begin{split}
    -\sum_{i} \Bigl( &\alpha x^{(\A)}_{i}(1 - x^{(\B)}_{i})^{\gamma}\log(x^{(\B)}_{i}) \\
    &+ (1 - \alpha)(1 - x^{(\A)}_{i})(x^{(\B)}_{i})^{\gamma}\log(1 - x^{(\B)}_{i}) \Bigr),
\end{split}
\label{eq:focal_loss}
\end{align}
where $\alpha$ is a balancing weight and the $\log$ arguments are clipped between $0.001$ and $0.999$.

Compared to the WBCE, the FL adds two factors $(1 - x^{(\B)}_{i})^{\gamma}$ and $(x^{(\B)}_{i})^{\gamma}$. However, while in the context of neural training, $x^{(\B)}_{i}$ is an occupancy probability, in the context of quality assessment, $x^{(\B)}_{i}$ is a binary value.
As a result, the FL is equivalent to the WBCE since $\gamma$ has no impact in the latter case. For this reason, we include the WBCE with $\alpha = 0.75$ in our experiments as an evaluation proxy for the FL used in \cite{quach_improved_2020}.

The neighborhood adaptive BCE (naBCE) \cite{guarda_neighborhood_2020} was proposed as an alternative to the BCE and FL \cite{lin_focal_2017}. It is a variant of the WBCE in which the weight $\alpha$ adapts to the neighborhood of each voxel $u$ resulting in a weight $\alpha_u$. 
Given a voxel $u$, its neighboorhood is a window $W$ of size $m \times m \times m$ centered on $u$.
Then, the neighborhood resemblance $r_u$ is the sum of the inverse euclidean distances of neighboring voxels with the same binary occupancy value as $u$.
Finally, the weight $\alpha_u$ is defined as $\alpha_u = \max(1 - r_u / \max(r)), 0.001)$ where $\max(r)$ is the maximum of all neighborhood resemblances.

The Mean Squared Error (MSE) on the TDF is expressed as follows
\begin{equation}
    \frac{1}{N} \sum_{i} (x^{(\A)}_{i} - x^{(\B)}_{i})^2.
\end{equation}

\subsection{Perceptual Loss}

We propose a perceptual loss based on differences between latent space representations learned by a neural network.
More precisely, we use an autoencoder as the underlying neural network. The model architecture of autoencoder used and its training procedure are presented in the following.

\subsubsection{Model architecture}

We adopt an autoencoder architecture based on 3D convolutions and transposed convolutions. Given an input voxel grid $x$, we perform an analysis transform $f_a(x) = y$ to obtain the latent space $y$ and a synthesis transform $f_s(y) = \tilde{x}$ as seen in Figure \ref{fig:perceptual_loss_model}. The analysis transform is composed of three convolutions with kernel size $5$, stride $2$ while the synthesis transform is composed of three transposed convolutions with same kernel size and stride. We use ReLU \cite{nair_rectified_2010} activations for all layers except for the last layer which uses a sigmoid activation.

\subsubsection{Training}

Using the previously defined architecture, we train two neural networks: one with explicit representation (binary) and another with implicit representation (TDF).

In the explicit case, we perform training of the perceptual loss with a focal loss function as defined in Eq. (\ref{eq:focal_loss}).
In the implicit case, we first define the Kronecker delta $\delta_i$ such that $\delta_i = 1$ when $i = 0$ and otherwise $\delta_i = 0$. Then, we define an adaptive MSE loss function for the training of the perceptual loss as follows
\begin{align}
\begin{split}
    \frac{1}{N} \sum_{i} \Bigl( &\delta_{1 - x^{(\A)}_{i}}w(x^{(\A)}_{i} - x^{(\B)}_{i})^2\\
    &+ (1 - \delta_{1 - x^{(\A)}_{i}})(1-w)(x^{(\A)}_{i} - x^{(\B)}_{i})^2 \Bigr),
\end{split}
\end{align}
where $w$ is a balancing weight. Specifically, we choose $w$ as the proportion of distances strictly inferior to $1$ with
\begin{equation}
w = \min\left(\max\left(\frac{\sum_{i} 1 - \delta_{1 - x^{(\A)}_{i}}}{N}, \beta\right), 1 - \beta\right).
\end{equation}
where $\beta$ is a bounding factor such that $w$ is bounded by $[\beta, 1 - \beta]$. This formulation compensates for class imbalance while avoiding extreme weight values.

In that way, the loss function adapts the contributions from the voxels that are far from occupied voxels ($x^{(\A)}_i = 1$) and voxels that are near occupied voxels ($x^{(\A)}_i < 1$). We train the network with the Adam \cite{kingma_adam:_2014} optimizer.

\begin{table}[t]
\centering
\caption{Table \ref{tab:results}: Statistical analysis of objective quality metrics.}
\label{tab:results}
\begin{tabular}{@{}lrrrr@{}} 
\toprule
Method     &           PCC &         SROCC &          RMSE &            OR \\ \midrule
TDF PL F9  &  $\bm{0.951}$ &  $\bm{0.947}$ &  $\bm{0.094}$ &  $\bm{0.375}$ \\ \hdashline
D2 MSE     &       $0.946$ &       $0.943$ &       $0.100$ &       $0.469$ \\ \hdashline
TDF MSE    &       $0.940$ &       $0.940$ &       $0.103$ &       $0.385$ \\ \hdashline
D1 MSE     &       $0.938$ &       $0.933$ &       $0.109$ &       $0.479$ \\ \hdashline
TDF PL     &       $0.935$ &       $0.933$ &       $0.110$ &       $0.490$ \\ \hdashline
Bin PL F1 &       $0.922$ &       $0.916$ &       $0.115$ &       $0.406$ \\ \hdashline
D2 PSNR    &       $0.900$ &       $0.898$ &       $0.129$ &       $0.500$ \\ \hdashline
Bin WBCE 0.75 &       $0.875$ &       $0.859$ &       $0.144$ &       $0.531$ \\ \hdashline
Bin PL    &       $0.863$ &       $0.867$ &       $0.151$ &       $0.552$ \\ \hdashline
D1 PSNR    &       $0.850$ &       $0.867$ &       $0.158$ &       $0.448$ \\ \hdashline
Bin naBCE &       $0.740$ &       $0.719$ &       $0.201$ &       $0.573$ \\ \hdashline
Bin BCE   &       $0.713$ &       $0.721$ &       $0.207$ &       $0.635$ \\ \hdashline
\bottomrule
\end{tabular}
\end{table} 

\subsubsection{Metric}

As seen in Figure \ref{fig:perceptual_loss_loss}, in order to compare two point clouds $x^{(\A)}$ and $x^{(\B)}$, we compute their respective latent spaces $f_a(x^{(\A)}) = y^{(\A)}$ and $f_a(x^{(\B)}) = y^{(\B)}$ using the previously trained analysis transform.
These latent spaces each have $F$ feature maps of size $W \times D \times H$ (width, depth, height).
Then, we define the MSE between latent spaces as follows
\begin{equation}
\frac{1}{N} \sum_i (y^{(\A)}_i - y^{(\B)}_i)^2.
\end{equation}
We compute this MSE either over all $F$ feature maps or on single feature maps.

\subsection{Point set metrics}

\subsubsection{Point-to-point (D1) and point-to-plane (D2)}

The point-to-point distance (D1) \cite{noauthor_common_2020} measures the average error between each point in $\A$ and their nearest neighbor in $\B$:
\begin{equation}
  e^{D1}_{\A,\B} = \frac{1}{N_{\A}} \sum_{\forall a_i \in \A} \lVert a_i - b_j \rVert^2_2
\end{equation}
where $b_j$ is the nearest neighbor of $a_i$ in $\B$.

In contrast to D1, the point-to-plane distance (D2) \cite{tian_geometric_2017} projects the error vector along the normal and is expressed as follows
\begin{equation}
  e^{D2}_{\A,\B} = \frac{1}{N_{\A}} \sum_{\forall a_i \in \A} ((a_i - b_j) \cdot n_i)^2
\end{equation}
where $b_j$ is the nearest neighbor of $a_i$ in $\B$ and $n_i$ is the normal vector at $a_i$.

The normals for original and distorted point clouds are computed with local quadric fittings using 9 nearest neighbors.

The D1 and D2 MSEs are the maximum of $e_{\A, \B}$ and $e_{\B, \A}$ and their Peak Signal-to-Noise Ratio (PSNR) is then computed with a peak error corresponding to three times the point cloud resolution as defined in \cite{noauthor_common_2020}.

\section{Experiments}

\subsection{Experimental setup}

We evaluate the metrics defined above on the ICIP2020 \cite{perry_quality_2020} subjective dataset. It contains 6 point clouds \cite{loop_microsoft_2016, deon_8i_2017} compressed using G-PCC Octree, G-PCC Trisoup and V-PCC with 5 different rates yielding a total of 96 stimuli (with 6 references) and their associated subjective scores.

\begin{figure*}[t]
\centering
\includegraphics[width=0.75\linewidth]{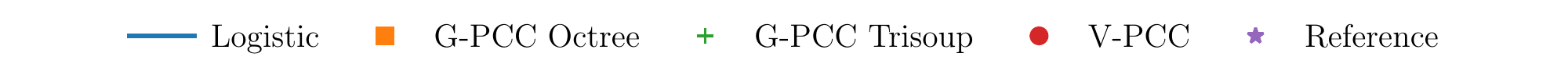}\\
\includegraphics[width=0.95\linewidth]{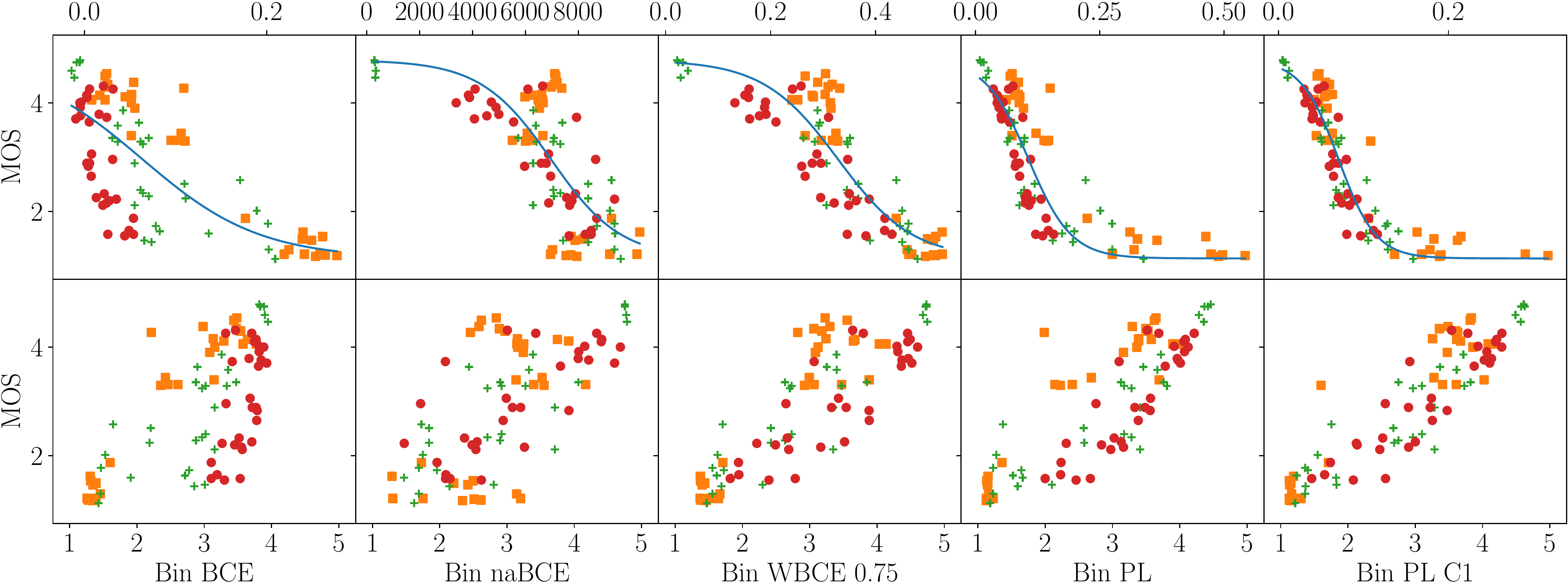}\\
\vspace{5pt}
\includegraphics[width=0.95\linewidth]{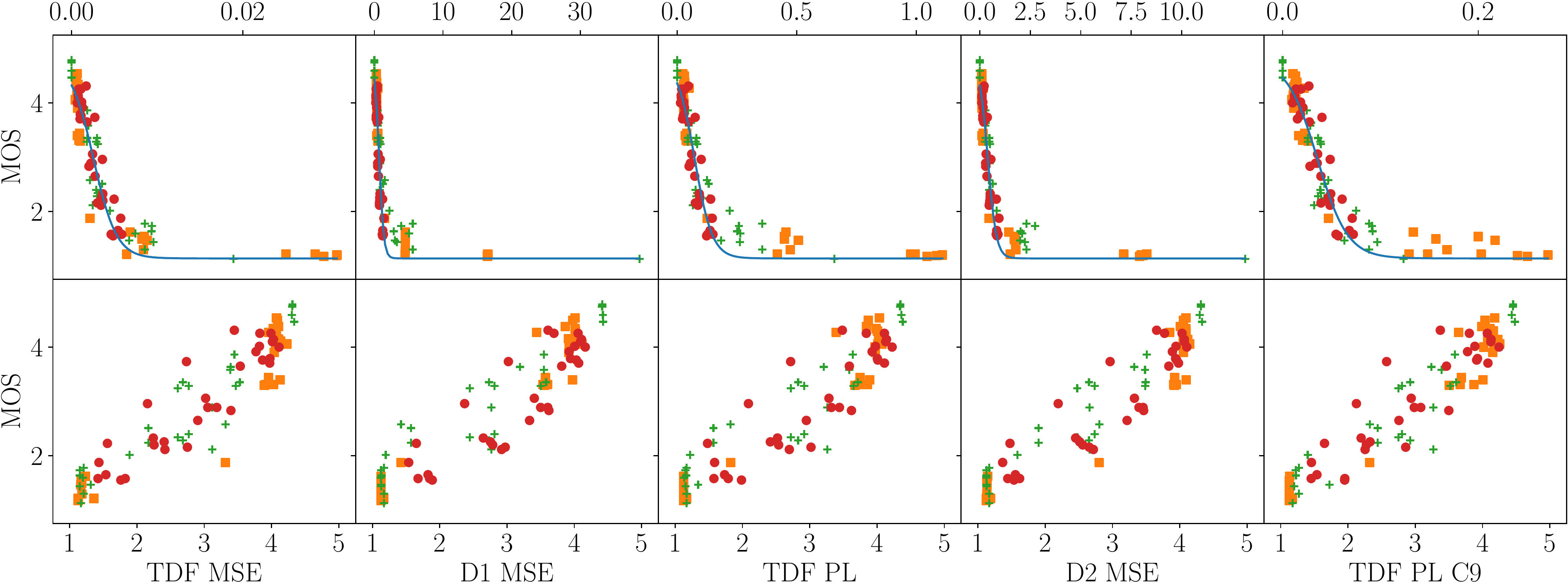}\\
\vspace{5pt}
\caption{Scatter plots between the objective quality metrics and the MOS values. The plots before and after logistic fitting are shown.}
\label{fig:scatter_plots}
\end{figure*}

For each metric, we compute the Pearson Correlation Coefficient (PCC), the Spearman Rank Order Correlation Coefficient (SROCC), the Root Mean Square Error (RMSE) and the Outlier Ratio (OR).
We evaluate the statistical significance of the differences between PCCs using the method in~\cite{zou_toward_2007}.
These metrics are computed after logistic fittings with cross-validation splits. Each split contains stimuli for one point cloud (i.e. reference point cloud and its distorted versions) as a test set and stimuli of all other point clouds as a training set.
The metrics are then computed after concatenating results for the test set of each split.
They are summarized in Table \ref{tab:metrics_table} and the values before and after logistic fitting are shown in Figure \ref{fig:scatter_plots}.

We use an upper bound value $u = 5$ when computing the TDF in Eq. (\ref{eq:tdf_trunc_norm}) and a block size of $64$ when block partitioning point clouds.
The naBCE window size is $m = 5$ as in the original paper.
The perceptual loss is trained with a learning rate of $0.001$, $\beta_1 = 0.9$ and $\beta_2 = 0.999$ on the ModelNet dataset \cite{sedaghat_orientation-boosted_2016} after block partitioning using Python 3.6.9 and TensorFlow \cite{abadi_tensorflow_2016} 1.15.0.

\subsection{Comparison of perceptual loss feature maps}

In our experiments, we first considered the perceptual loss computed over all feature maps.
However, we observed that some feature maps are more perceptually relevant than others.
Consequently, we include the best feature maps for each voxel grid representation in our results.
This corresponds to feature map 9 (TDF PL F9) for TDF PL and 1 (Bin PL F1) for Bin PL.

Moreover, We observe that some feature maps are unused by the neural network (constant).
Therefore, they exhibit high RMSE values (all equal to $0.812$) as their perceptual loss MSE are equal to $0$. 
Specifically, we observe that TDF PL has 6 unused feature maps, while Bin PL has a single unused feature map.
This suggests that the perceptual loss learns a sparser latent space representation when using TDF compared to binary.
Thus, implicit representations may improve compression performance compared to explicit representations as fewer feature maps may be needed.

\subsection{Comparison of objective quality metrics}

In Table \ref{tab:results}, we observe that the TDF PL F9 is the best method overall. In particular, identifying the most perceptually relevant feature map and computing the MSE on this feature map provides a significant improvement. Specifically, the difference between the PCCs of TDF PL F9 and TDF PL is statistically significant with a confidence of $95\%$.

For voxel grid metrics, we observe that TDF metrics perform better than binary metrics.
In particular, the RMSEs of the former are noticeably lower for point clouds compressed with G-PCC Octree compared to the RMSE of the latter as can be seen in Table \ref{tab:results_by_compression}.
This suggests that implicit representations may be better at dealing with density differences between point clouds in the context of point cloud quality assessment.

\begin{table*}[h!]
\centering
\caption{Table \ref{tab:results_by_compression}: Statistical analysis of objective quality metrics by compression method.}
\label{tab:results_by_compression}
\begin{tabular}{@{}lcrrrcrrrcrrr@{}}
\toprule
\multirow{2}{*}{Method} && \multicolumn{3}{c}{G-PCC Octree} && \multicolumn{3}{c}{G-PCC Trisoup} && \multicolumn{3}{c}{V-PCC} \\
\cmidrule{3-5} \cmidrule{7-9} \cmidrule{11-13}
            &&            PCC &         SROCC &          RMSE &&           PCC &         SROCC &          RMSE &&           PCC &         SROCC &          RMSE \\ \midrule[\heavyrulewidth]
TDF PL F9   &&  $\mathit{0.975}$ & $\mathit{0.859}$ &  $\bm{0.078}$ &&       $0.936$ &       $0.910$ &  $\bm{0.101}$ &&       $0.897$ &       $0.850$ &       $0.106$ \\ \hdashline
D2 MSE      &&        $0.962$ &       $0.829$ &       $0.094$ &&  $\bm{0.954}$ &  $\bm{0.924}$ & $\mathit{0.103}$ && $\mathit{0.903}$ &       $0.860$ & $\mathit{0.103}$ \\ \hdashline
TDF MSE     &&        $0.952$ &       $0.839$ &       $0.106$ &&       $0.933$ &       $0.917$ &       $0.106$ &&  $\bm{0.912}$ &  $\bm{0.867}$ &  $\bm{0.098}$ \\ \hdashline
D1 MSE      &&   $\bm{0.976}$ &       $0.851$ & $\mathit{0.082}$ && $\mathit{0.937}$ & $\mathit{0.918}$ &       $0.126$ &&       $0.876$ &       $0.844$ &       $0.119$ \\ \hdashline
TDF PL      &&        $0.970$ &       $0.840$ &       $0.087$ &&       $0.918$ &       $0.900$ &       $0.127$ &&       $0.876$ &       $0.837$ &       $0.115$ \\ \hdashline
Bin PL F1  &&        $0.941$ &       $0.786$ &       $0.138$ &&       $0.927$ &       $0.907$ &       $0.107$ &&       $0.898$ & $\mathit{0.865}$ &       $0.109$ \\ \hdashline
D2 PSNR     &&        $0.943$ &  $\bm{0.890}$ &       $0.110$ &&       $0.926$ &       $0.895$ &       $0.108$ &&       $0.738$ &       $0.723$ &       $0.166$ \\ \hdashline
Bin WBCE 0.75  &&        $0.923$ &       $0.747$ &       $0.163$ &&       $0.918$ &       $0.886$ &       $0.112$ &&       $0.850$ &       $0.786$ &       $0.164$ \\ \hdashline
Bin PL     &&        $0.931$ &       $0.852$ &       $0.186$ &&       $0.892$ &       $0.886$ &       $0.130$ &&       $0.880$ &       $0.852$ &       $0.142$ \\ \hdashline
D1 PSNR     &&        $0.903$ &       $0.859$ &       $0.156$ &&       $0.910$ &       $0.895$ &       $0.117$ &&       $0.599$ &       $0.689$ &       $0.202$ \\ \hdashline
Bin naBCE  &&        $0.552$ &       $0.357$ &       $0.277$ &&       $0.846$ &       $0.786$ &       $0.154$ &&       $0.748$ &       $0.692$ &       $0.170$ \\ \hdashline
Bin BCE    &&        $0.946$ &       $0.841$ &       $0.188$ &&       $0.776$ &       $0.800$ &       $0.177$ &&       $0.574$ &       $0.500$ &       $0.250$ \\ \hdashline
\bottomrule
\end{tabular}%
\end{table*}

\section{Conclusion}

We proposed a novel perceptual loss that outperforms existing objective quality metrics and is differentiable in the voxel grid.
As a result, it can be used as a loss function in deep neural networks for point cloud compression and it is more correlated with perceived visual quality compared to traditional loss functions such as the BCE and the focal loss.
Overall, metrics on the proposed implicit TDF representation performed better than explicit binary representation metrics. Additionally, we observed that the TDF representation yields sparser latent space representations compared to the binary representations. 
This suggests that switching from binary to the TDF representation may improve compression performance in addition to enabling the use of better loss functions.

\section{Acknowledgments} 

We would like to thank the authors of \cite{guarda_neighborhood_2020} for providing their implementation of naBCE.
This work was funded by the ANR ReVeRy national fund (REVERY ANR-17-CE23-0020).

\small

\normalsize
\begin{biography}
Maurice Quach received the Computer Science Engineer Diploma from University of Technology of Compi\`egne in 2018. He is currently studying for a PhD on Point
Cloud Compression and Quality Assessment under the supervision of Frederic Dufaux and Giuseppe Valenzise at Universit\'e Paris-Saclay, CNRS, CentraleSup\'elec, Laboratoire des signaux et syst\`emes, France.
\end{biography}

\end{document}